\documentclass[letterpaper, 10 pt, conference]{ieeeconf}  

\IEEEoverridecommandlockouts                              

\title{\LARGE \bf
MAG-Nav: Language-Driven Object Navigation Leveraging Memory-Reserved Active Grounding
}

\author{Weifan Zhang$^{1,2*}$, Tingguang Li$^{1*}$, Yuzhen Liu$^{1\dag}$%
\thanks{* Equal contribution.}%
\thanks{$\dag$ Corresponding author.}%
\thanks{$^{1}$ Tencent Robotics X, Shenzhen, China.}%
\thanks{$^{2}$ Harbin Institute of Technology, Shenzhen, China.}%
}

\usepackage{booktabs}  
\usepackage{colortbl}  
\usepackage{xcolor}    
\usepackage{multirow} 
\usepackage{siunitx}
\usepackage{cite}
\usepackage{mathptmx}
\usepackage{amsmath}
\usepackage{graphicx}
\usepackage{amsfonts}  
\usepackage{subcaption}
\usepackage{algorithm}
\usepackage{algorithmic}

\usepackage{color}

\begin{document}

\maketitle
\thispagestyle{empty}
\pagestyle{empty}

\begin{figure*}[t]
  \centering
  \includegraphics[width=0.9\textwidth]{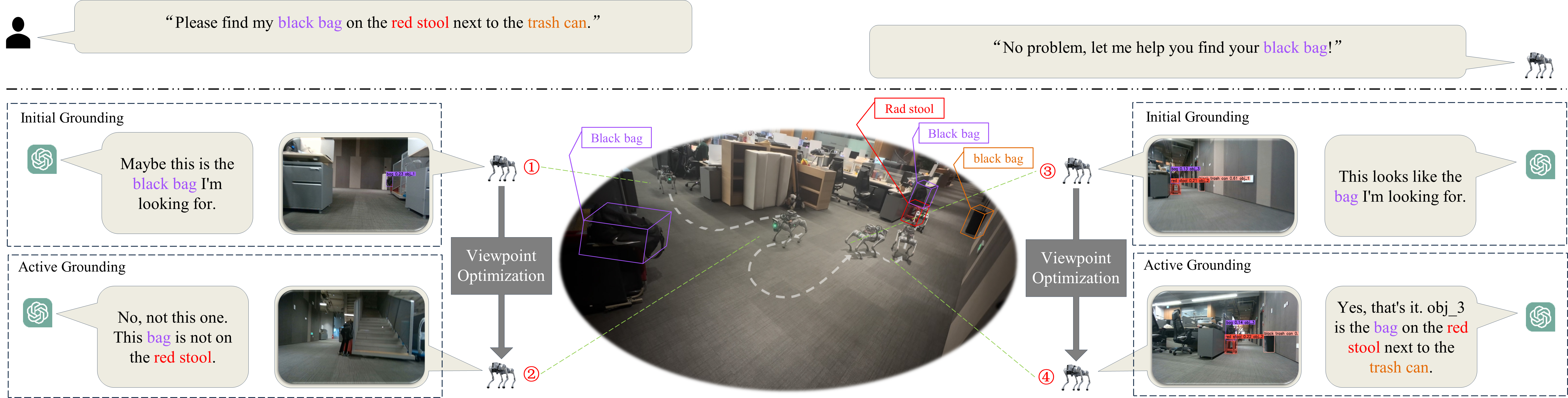}
  \caption{
  Example of active grounding. Given the instruction "Please find my black bag on the red stool next to the trash can," the robot initially misidentifies a distractor bag on a folding bed due to a limited viewpoint. After view planning, it correctly rejects the distractor, then discovers the true target and confirms it via active grounding before navigating to it.
  }
  \label{fig:real_world_overview}
\end{figure*}

\begin{abstract}
Visual navigation in unknown environments based solely on natural language descriptions is a key capability for intelligent robots. In this work, we propose a navigation framework built upon off-the-shelf Visual Language Models (VLMs), enhanced with two human-inspired mechanisms: perspective-based active grounding, which dynamically adjusts the robot’s viewpoint for improved visual inspection, and historical memory backtracking, which enables the system to retain and re-evaluate uncertain observations over time.
Unlike existing approaches that passively rely on incidental visual inputs, our method actively optimizes perception and leverages memory to resolve ambiguity, significantly improving vision-language grounding in complex, unseen environments. Our framework operates in a zero-shot manner, achieving strong generalization to diverse and open-ended language descriptions without requiring labeled data or model fine-tuning.
Experimental results on Habitat-Matterport 3D (HM3D) show that our method outperforms state-of-the-art approaches in language-driven object navigation. We further demonstrate its practicality through real-world deployment on a quadruped robot, achieving robust and effective navigation performance.
\end{abstract}

\section{Introduction}

Embodied intelligent robots with advanced autonomous navigation capabilities are expected to understand and interact with real-world environments based on natural human instructions. 
However, recent progress in robot vision-and-language navigation (VLN) often imposes notable limitations on the scope of such instructions. 
Many approaches restrict interaction to incremental path descriptions that rely heavily on prior environmental knowledge \cite{vasudevan2021talk2nav, shah2023lm, huang2023visual}, while others confine object references to a limited set of predefined categories \cite{Chaplot_Gandhi_Gupta_Salakhutdinov_2020, yu2023l3mvn, yokoyama2024vlfm}. These constraints limit the flexibility and generalization of embodied agents.

In contrast to these constrained settings, everyday life involves issuing open-ended, context-rich instructions about objects. For instance, one might ask a robotic assistant: "Please confirm whether my black comb is on the coffee table next to the sofa." This instruction includes both spatial context (on the coffee table next to the sofa) and physical attributes (black comb), reflecting the richness of realistic human-robot interactions. Addressing this gap motivates our focus on zero-shot language object navigation, where a robot must locate and navigate to an object described in free-form language within a previously unseen environment.

A key challenge in this task lies in effectively linking language comprehension with visual perception to accurately ground the target object in the scene. Recent advances in Visual Language Models (VLMs), such as CLIP \cite{radford2021learning} and GPT-4O \cite{yang2024llm}, have shown promise in bridging this modality gap. For example, the CoW algorithm \cite{gadre2023cows} leverages CLIP to divide navigation into exploration for ambiguous targets and planning for well-defined ones. However, CLIP-based models often struggle with complex prompts due to their "bag-of-words" nature \cite{yuksekgonul2022and}, leading to contextual misinterpretations.

Despite the impressive progress in Vision Language Models (VLMs), their application to robotic navigation remains largely passive—relying on whatever visual input is incidentally captured during exploration. This approach is fundamentally limited by the nature of how VLMs are trained: on large-scale, static, third-person image-caption datasets that lack the embodied and sequential nature of robot perception. As a result, standard VLMs often fail in dynamic, egocentric settings due to occlusions, suboptimal viewpoints, or ambiguous object appearances.
More critically, when multiple objects match the language description, decisions based solely on current observations can lead to premature or incorrect grounding. To overcome these limitations, we argue that robots must go beyond passive perception and instead adopt two key capabilities inspired by human cognition: active viewpoint selection, which enables more informative visual inspection, and contextual memory recall, which allows revisiting uncertain observations over time.

We propose a framework that integrates these mechanisms into robotic navigation through an embodied interaction loop with off-the-shelf VLMs. Our method begins by constructing visual-spatial object hypotheses using a perception module. It then employs a perspective-based active grounding strategy, where the robot dynamically adjusts its viewpoint to maximize the discriminability of candidate objects. This process compensates for the mismatch between VLM training data and the first-person, dynamic nature of robotic vision.
In parallel, we introduce a historical memory backtracking mechanism that emulates human-like memory recall. Uncertain observations are selectively stored and re-evaluated in light of new context, enabling the system to resolve ambiguity across time ("was that the target or just a similar object?") and transfer knowledge between tasks by cross-referencing previously seen objects.

To our knowledge, this work presents the first successful integration of off-the-shelf VLMs with embodied active perception and contextual memory reasoning. Through simulation and real-world validation, we demonstrate that this combination significantly improves navigation accuracy and robustness in complex environments, as shown in Fig. \ref{fig:real_world_overview}.

In summary, our main contributions are as follows:

\begin{itemize}
\item We propose a perspective-based active grounding method that enables robots to actively reposition themselves for more informative visual observations, thereby enhancing VLM performance by improving the quality of input data—mirroring how humans adjust their viewpoints for clearer perception.

\item We introduce a historical memory backtracking mechanism that emulates human-like recall, allowing the system to revisit and re-evaluate uncertain observations over time. This enables context-aware decision-making and cross-task knowledge transfer in complex environments.

\item We provide extensive experimental validation, including simulation benchmarks (e.g., HM3D) and real-world deployment on a quadruped robot, demonstrating the robustness, generalization, and practical effectiveness of our approach.

\item We publicly release our code to facilitate future research and promote reproducibility in embodied vision-language navigation.
\end{itemize}

\section{Related Work}

\textbf{Object Navigation. }
Object navigation is a critical task for intelligent robots, where an agent is required to navigate to a specified location based on point goals \cite{Wijmans_Kadian_Morcos_Lee_Essa_Parikh_Savva_Batra_2019, Chaplot_Gandhi_Gupta_Gupta_Salakhutdinov_2020, Chattopadhyay_Hoffman_Mottaghi_Kembhavi_2021, Gordon_Kadian_Parikh_Hoffman_Batra_2019, Hahn_Chaplot_Tulsiani_Mukadam_Rehg_Gupta_2021, Savva_Chang_Dosovitskiy_Funkhouser_Koltun_2017, Xia_Zamir_He_Sax_Malik_Savarese_2018, Yan_Misra_Bennett_Walsman_Bisk_Artzi_2018}, image goals \cite{Mezghan_Sukhbaatar_Lavril_Maksymets_Batra_Bojanowski_Alahari_2022, ramakrishnan2022poni, Zhu_Mottaghi_Kolve_Lim_Gupta_Fei-Fei_Farhadi_2017}, or object goals \cite{al2022zero, Chang_Gupta_Gupta_2020, Chaplot_Gandhi_Gupta_Salakhutdinov_2020, Chattopadhyay_Hoffman_Mottaghi_Kembhavi_2021, Liang_Chen_Song_2021, Wani_Patel_Jain_Chang_Savva_2020, Wortsman_Ehsani_Rastegari_Farhadi_Mottaghi_2019, Yang_Wang_Farhadi_Gupta_Mottaghi_2018}. In this work, we focus on object goal navigation tasks within the VLM framework. However, existing object goal navigation tasks often directly use the object's text name as input to the navigation framework, overlooking the diverse semantic structures that composite text input can provide (such as spatial information, physical properties, and similar objects). In this paper, we aim to empower the agent with the ability to understand complex language goal descriptions and navigate towards language goals in unknown environments using VLM, without providing detailed step-by-step instructions.

\textbf{Zero-Shot Object Navigation. }
Zero-Shot Object Navigation \cite{gadre2023cows} extends the ObjectNav task to an open-vocabulary setting, evaluating the agent on objects it has not been explicitly trained on. Achieving open-vocabulary object goal navigation is a more realistic scenario, with some methods \cite{khandelwal2022simple, majumdar2022zson} training the agent on a subset of categories and then evaluating it on unseen objects. Majumdar et al. \cite{majumdar2022zson} encodes the goal image into a multimodal semantic embedding space, training the robot to perform semantic goal navigation in unannotated 3D environments. However, these methods still require extensive simulated training, limiting their scalability. Recent trends involve leveraging Large Language Models (LLMs) for unsupervised zero-shot goal navigation \cite{zhou2023esc, yu2023l3mvn, wu2024voronav, shah2023navigation}. These methods leverage the common-sense knowledge embedded in Large Language Models (LLMs) to interactively navigate to the goal, demonstrating strong generalization capabilities in adapting to new tasks and environments without additional training.

\textbf{Visual Grounding. }
The process of visual grounding entails associating particular regions of an image with natural language descriptions, which constitutes a pivotal element in the domain of robot vision and language navigation. Conventional object detection models, which are designed to identify objects belonging to predefined categories, often exhibit limitations in their ability to generalise to new classes. Inspired by advancements in open-vocabulary segmentation, researchers have explored open-vocabulary grounding. However, these methods often rely heavily on models like CLIP \cite{radford2021learning}, which exhibit a "bag-of-words" \cite{yuksekgonul2022and} behavior and lack compositional understanding. In contrast, our work introduces a new approach that leverages the capabilities of Large Language Models (LLMs) for 2D visual grounding tasks. In contrast to models that have been optimised for specific tasks \cite{wang2023cogvlm, zhang2023llava, ha2022semantic, hong20233d}, our model is designed to be flexible, capable of interpreting diverse semantic structures in the query, such as spatial information, physical attributes, and distinctions among multiple objects of the same class. This expands the scope and effectiveness of visual grounding in different scenarios, achieving a more universal application. Furthermore, our approach operates without any additional training or fine-tuning, demonstrating the practical value and efficiency of LLMs and LMMs in visual grounding tasks.

\section{PROBLEM FORMULATION}

\begin{figure*}[t]
  \centering
  \includegraphics[width=0.9\textwidth]{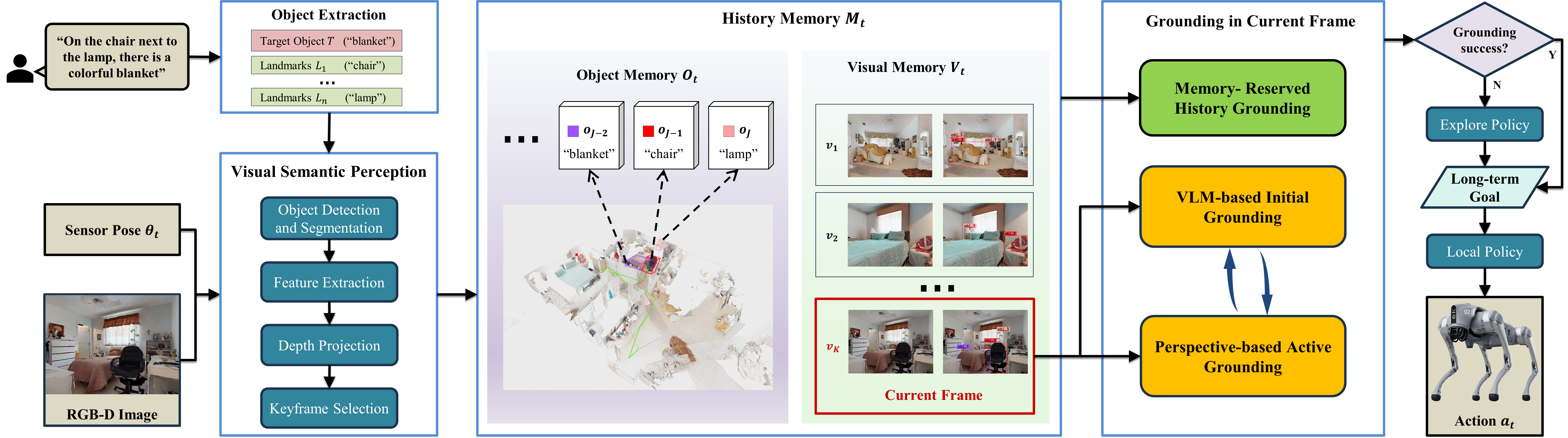}
  \caption{
  Overview of the proposed MAG-Nav system. Given a natural language instruction, the system extracts the target category and related landmarks via a large language model (LLM). A visual semantic perception module incrementally builds object and visual memories from egocentric observations. The memory-augmented active grounding framework integrates view planning and memory replay to continuously refine object grounding. Based on grounding status, the robot either navigates directly to the object or performs frontier-based exploration. A hierarchical motion policy ensures efficient and safe goal reaching.
  }
  \label{fig:system_overview}
\end{figure*}

We address the task of language-driven zero-shot object navigation (L-ZSON)\cite{gadre2023cows}, where a robot must find a target object based on a natural language description.
Compared with traditional ObjectNav tasks\cite{batra2020objectnav}, L-ZSON is more challenging yet applicable because it requires the robot to understand high-level language descriptions that contain different levels of granularity, such as spatial relationships (e.g., "Find the water cup in front of the TV") and physical attributes (e.g., "Go find the red cabinet"), rather than relying solely on object categories. 
Regarding the sensor setup, the robot is equipped with an egocentric RGB-D camera and an odometry sensor that provides its current forward and horizontal distances and heading relative to its starting pose.

\section{Method}

\subsection{System Overview}

Fig. \ref{fig:system_overview} illustrates the proposed system framework. The system first leverages a Large Language Model (LLM) to extract the target object's category name (denoted as  \(T \)) and identify relevant landmark categories (\(L = \{L_1, L_2, \ldots, L_n\} \))  from the user's natural language instruction. Concurrently, a proposed visual semantic perception module processes the robot's visual input to incrementally construct both visual and object memories.
A key innovation of our approach is the memory-augmented active grounding framework, which overcomes the limitations of single-stage grounding by dynamically integrating view planning and memory replay mechanisms.
Built upon this framework, our navigation strategy adapts dynamically to the grounding results: when an object is successfully localized, it becomes the target for long-term navigation; otherwise, the system autonomously shifts to frontier-based exploration to identify high-potential unexplored regions. In both cases, a hierarchical motion policy ensures efficient and collision-free goal-reaching.

\subsection{Visual Semantic Perception Module}

In the task of robot vision-language navigation, several studies\cite{gu2024conceptgraphs,shah2023lm} maintain historical memory to help the agent better understand the environment and tasks, thereby making more informed decisions.
However, these studies primarily focus on constructing 3D topological graphs of object instances, neglecting the wealth of information that can be derived from visual observations of objects.

In our framework, we not only maintain the memory of 3D object instances but also preserve keyframe visual memory. In addition, we establish a unique mapping from this keyframe visual memory to 3D object instances in order to locate the position of objects displayed in the image within the 3D space.

\subsubsection{Object Memory and Visual Memory}

Given a sequence of RGB-D observations \( I = \{I_1, I_2, . . . , I_t\} \), we construct historical memory \( M_t = \langle O_t, V_t \rangle \), where \( O_t = \{\mathbf{o_j}\}_{j=1}^{J} \) and \( V_t = \{\mathbf{v_k}\}_{k=1}^{K} \) represent the sets of objects and image pairs, respectively.
Each object \( \mathbf{o_j} \) is characterized by a 3D point cloud \( p_{o_j} \) and a semantic feature vector \( f_{o_j} \). This 3D object memory is updated in real-time, incorporating each incoming frame \( I_t = \langle I^t_{\text{rgb}} , I^t_{\text{depth}} , \theta_t \rangle \) (color image, depth image, pose) into the existing object set \( O_{t-1} \), by either adding to existing objects or instantiating new ones.
Each image pair \( \mathbf{v_k} \) is characterized by an original RGB image \( I^t_{\text{rgb}} \) and an annotated image \( I^t_{\text{annot}} \). We build this visual memory incrementally, adding the observations deemed as keyframes to the existing visual memory \( V_{t-1} \).

\subsubsection{Visual Memory Unit}

Based on the current frame observation \( I_t \), we construct a visual memory unit \( \mathbf{v_t} = \langle I^t_{\text{rgb}}, I^t_{\text{annot}} \rangle \).
In this definition, \( I^t_{\text{annot}} \) is the annotated RGB image, where detected targets and landmarks are marked with bounding boxes, each labeled with a unique identifier.
To obtain the bounding boxes of targets and landmarks, we deploy a pre-trained, open-vocabulary object detection model.
Specifically, we take the target and landmarks class name as textual prompts, and utilise a open-vocabulary detection model to predict the bounding box in the current frame RGB image \( I^t_{\text{rgb}} \).
Once the bounding boxes are detected, we generate an annotated image \( I^t_{\text{annot}} \) by labeling each bounding box with a unique identifier from 1 to \(N_{\text{object}}\) at its top-left corner.
This annotated image will provide a crucial visual prompt for subsequent visual-language grounding.

\subsubsection{Object Memory Unit}

To obtain the complete 3D information of each object \(\mathbf{o^t_{i}}\) detected in \( I^t_{\text{annot}} \), a class-agnostic segmentation model is used to obtain a set of masks \(\{m_{t,i}\}_{i=1}^{N_{\text{object}}}\) corresponding to these objects.
Each extracted mask \(m_{t,i}\) is then passed to a visual feature extractor (CLIP) to obtain a visual descriptor \(f_{t,i}\).
We alse use the depth value of each pixel within the mask to compute its 3D coordinates in the camera coordinate frame and these 3D points are then transformed to the world coordinate frame.
This results in a point cloud \(p_{t,i}\) and its corresponding unit-normalized semantic feature vector \(f_{t,i}\).

\subsubsection{Keyframe Selection and Memory Update}

We determine a visual memory Unit as a keyframe if its annotated image introduces a new target object.
For each object \(\mathbf{o^t_{i}} = \langle p_{t,i}, f_{t,i} \rangle \) detected in \( I^t_{\text{annot}} \), we assess whether it is a new one by comparing its similarity to objects in the object memory \( O_{t-1} \). This involves calculating both visual and spatial similarities and combining them into an overall similarity score, as per the method in \cite{gu2024conceptgraphs}. 
If the similarity score surpasses a certain threshold, \(\delta_{\text{sim}}\), we associate the detected object with the most similar one in \( O_{t-1} \), subsequently updating \( O_{t} \).
On the other hand, if no object in \( O_{t-1} \) has a similarity score higher than \(\delta_{\text{sim}}\) with the detected object, we regard this detected object as a new entity and incorporate it into \( O_{t} \). At the same time, we mark the current frame \( \mathbf{v_t} \) as a keyframe and add it to the visual memory \( V_{t-1} \) for future grounding.

\subsection{Memory-Reserved Active Grounding}

In this section, we propose a method to accurately ground the object described in the language instruction within the visual observation and determine its spatial coordinates using object memory.

Numerous studies \cite{gadre2023cows, yokoyama2024vlfm} have underscored the advantages of using Vision-and-Language Models (VLMs) for grounding tasks. These models effectively merge verbal descriptions with visual data, thereby enhancing an agent's ability to discern spatial relationships and object attributes.

However, in terms of grounding methods, \cite{yang2024llm} converts visual observations into language descriptions and uses a large language model (LLM) for grounding, thereby losing a significant amount of auxiliary information implicit in the original visual observations. \cite{gadre2023cows} uses discriminative VLMs like CLIP to directly score the similarity between language descriptions and visual observations, lacking a profound understanding of the text and images. Conversely, methods that directly feed visual prompts into generative VLMs for grounding have not been widely explored in navigation tasks.

Moreover, existing methods \cite{yang2024llm, gadre2023cows, gu2024conceptgraphs, yokoyama2024vlfm} typically ground objects passively based on random viewpoints encountered during exploration, which may lead to unsatisfactory grounding results. This is because the viewpoints may not provide a complete or appropriate view of the target object due to occlusion of vision, inappropriate distance, etc.

Lastly, these grounding methods mostly determine the grounding target based on the current observation, providing a binary grounding result. However, in many scenarios, the agent is confronted with ambiguous targets. Therefore, it is crucial to store uncertain key observations and revisit these observations after the exploration phase.

To address these limitations, we propose the Memory-Reserved Active Grounding method. This method leverages the understanding capabilities of generative VLMs and the mobility of robots, dividing grounding into three stages: VLM-based initial grounding, perspective-based active grounding, and memory-based reserved grounding (see Algorithm \ref{alg:grounding}).

In the initial grounding stage, the VLM uses visual and language prompts to ground the target object under the keyframe perspective during the robot's exploration process. While this phase is crucial for initial object recognition, it may not always provide the most appropriate viewpoint of the target object. The active grounding phase aims to overcome this challenge.
Once the VLM identifies a specific object as a target during the initial grounding phase, we invoke a viewpoint planning algorithm to optimize the best viewpoint for that target, which takes into account the target object's line-of-sight occlusion, viewing distance, and completeness. The robot then actively navigates to this viewpoint and performs a second round of VLM grounding.
Only after both rounds of grounding are successful, the target is considered to be accurately identified and localized, and thus the robot is activated to navigate to the target location. Otherwise, we activate the exploration module to continue the search. If the target cannot be determined by the end of the exploration, we initiate the memory-based reserved grounding, selecting a target from historical memory for grounding.



\begin{algorithm}[tb]
  \caption{Memory-Reserved Active Grounding}
  \label{alg:grounding}
\textbf{Input}: Target description $L_\text{text}$ \\
\textbf{Output}: Target $\mathbf{P}_{goal}$
\begin{algorithmic}[1]
\STATE Initialize variables: $\mathbf{P}_{obj}, \mathbf{P}_{view}, \mathbf{P}_{explore}$
\STATE Initialize Memory: $M_t \leftarrow \text{None}$
\STATE $found \leftarrow \text{False}$
\WHILE{not $found$}
    \STATE $v_t \leftarrow \text{getObservation}()$
    \IF{isExploring()}
        \IF{$v_t$ is keyframe}
            \STATE $result \leftarrow \text{VLMGrounding}(v_t, L_\text{text})$
            \STATE $M_t \leftarrow \text{updateMemory}(result, v_t, M_t)$
            \IF{$result.success$}
                \STATE $\mathbf{P}_{obj} \leftarrow \text{index3DPos}(v_t, result)$
                \STATE $\mathbf{P}_{view} \leftarrow \text{optimizeViewpoint}(\mathbf{P}_{obj})$
                \STATE Navigate to $\mathbf{P}_{view}$ and get new image $v_t$
                \STATE $result \leftarrow \text{ActiveGrounding}(v_t, L_\text{text})$
                \IF{$result.success$}
                    \STATE $\mathbf{P}_{goal} \leftarrow \text{index3DPos}(v_t, result)$
                    \STATE $found \leftarrow \text{True}$
                \ENDIF
            \ENDIF
        \ENDIF
        \IF{not $found$}
            \STATE $\mathbf{P}_{explore} \leftarrow \text{exploreEnv}()$
            \STATE Navigate to $\mathbf{P}_{explore}$
        \ENDIF
    \ELSE
        \STATE $\mathbf{P}_{goal} \leftarrow \text{reservedGrounding}(M_t, L_\text{text})$
        \STATE $found \leftarrow \text{True}$
    \ENDIF
\ENDWHILE
\STATE \textbf{return} $\mathbf{P}_{goal}$
\end{algorithmic}
\end{algorithm}

\subsubsection{VLM-based Grounding using Visual Prompts}

The annotated image \(I^t_{\text{annot}}\) and original RGB image \(I^t_{\text{rgb}}\) of current visual memory unit \( \mathbf{v_t} \) are used as visual prompts, while the linguistic description of the target object \( \mathbf{L}_\text{text} \) is used as textual prompt. The annotated image \(I^t_{\text{annot}}\) contains the bounding boxes with their respective identifiers, while \(I^t_{\text{rgb}}\) provides the original visual context information. We found that the inclusion of raw images \(I^t_{\text{rgb}}\) is essential in the VLM grounding process, which will be discussed in the ablation study section.
These inputs are fed into the VLM to determine which of the annotated candidates most closely matches the target described in the text. 
The output format of the VLM is predefined to be the numerical identifier \(i\) (\(i = 1 \ldots N_{\text{object}}\)) of the bounding box that best matches the textual description provided. 
This identifier is used to index the previously saved bounding boxes, thereby pinpointing the exact location of the target object \(\mathbf{o^t_{i}} \).

\subsubsection{Active Grounding based-on Viewpoint Optimization}
In the initial grounding phase, the robot's viewpoint of the target is limited to the first observation position where the target is seen during exploration. However, this position is random and not optimal for the target, and therefore may not provide complete information about the target and landmarks, which in turn affects the quality of VLM grounding. To address this limitation, we propose a viewpoint optimization algorithm to optimize the best observation position about the target \( d^T_i \), allowing the robot to autonomously navigate to that position for active grounding.

The goal of the viewpoint optimization problem is to find an optimal viewpoint such that the target object can be best observed from that viewpoint.
The viewpoint, the optimization variable of this problem, is typically defined in the SE(3) space: \( \mathbf{v} \in \mathrm{SE}(3) \). However, in our task, the viewpoint is bound to the pose of the ground robot, so we can simplify the viewpoint to a point in the SE(2) space: \( \mathbf{v} \in \mathrm{SE}(2) \). Additionally, since there is only one observed target, the robot's pose at different positions must always point towards the target, which can be considered as a constraint. Therefore, the optimization variable can be further simplified to a point in a 2D grid map: \( \mathbf{v} \in \mathbb{Z}^2 \).
Thus, the viewpoint optimization problem can be defined as a multi-objective discrete optimization problem, where the objective function \( f(\mathbf{v}) \) is defined as:
\[
f(\mathbf{v}) = -R_{\text{visible}}(\mathbf{v}) - R_{\text{fov}}(\mathbf{v}) + P_{\text{distance}}(\mathbf{v}) + P_{\text{feasibility}}(\mathbf{v})
\]
where \( R_{\text{visible}}(\mathbf{v}) \) denotes the visibility reward, which ensures that the viewpoint and the target are not blocked by other objects. \( R_{\text{fov}}(\mathbf{v}) \) accounts for the field of view, aiming to maximize the size of the target within the observer's field of view. Meanwhile, \( P_{\text{distance}}(\mathbf{v}) \) represents the distance penalty, which is used to optimize the distance from the observation point to the target so that it is neither too close (resulting in the inability to fully observe the target) nor too far (resulting in unclear observation of the target). \( P_{\text{feasibility}}(\mathbf{v}) \) represents the feasibility penalty, ensuring that the observation location is a physically traversable voint.

The visibility reward \( R_{\text{visible}}(\mathbf{v}) \) is defined as:
\[
R_{\text{visible}}(\mathbf{v}) = w_{\text{visible}} \cdot \min(N_{\text{visible}}(\mathbf{v}, G), 3)
\]
where \( w_{\text{visible}} \) is the visibility weight, and \( G = \{g_1, g_2, g_3, g_4\} \) represents the set of four boundary points of the target object's point cloud projected onto the 2D grid map. \( N_{\text{visible}}(\mathbf{v}, G) \) is the number of target points visible from \(\mathbf{v}\).

The field of view reward \( R_{\text{fov}}(\mathbf{v}) \) is defined as:
\[
R_{\text{fov}}(\mathbf{v}) = 
\begin{cases} 
w_{\text{fov}} \cdot \theta_{\text{max}}(\mathbf{v}) & \text{if } \theta_{\text{max}}(\mathbf{v}) < \text{fov} \\ 
0 & \text{otherwise}
\end{cases}
\]
where \( w_{\text{fov}} \) is the field of view weight, and \(\theta_{\text{max}}(\mathbf{v})\) is the maximum angular separation between the viewpoint and all target points.

The distance penalty \( P_{\text{distance}}(\mathbf{v}) \) is defined as:
\[
P_{\text{distance}}(\mathbf{v}) = w_{\text{distance}} \cdot \frac{1}{|G|} \sum_{g \in G} \left| (\|\mathbf{v} - g\| - d_{\text{desired}}) \right|
\]
where \( w_{\text{distance}} \) is the weight for the distance penalty and \( d_{\text{desired}} \) is the desired distance.

The feasibility penalty \( P_{\text{feasibility}}(\mathbf{v}) \) is defined as:
\[
P_{\text{feasibility}}(\mathbf{v}) = \begin{cases} 
C_{\text{infeasible}} & \text{if obstacles or unexplored} \\ 
0 & \text{otherwise}
\end{cases}
\]
where \( C_{\text{infeasible}} \) is a large constant used to penalize viewpoints located in obstacle or unexplored areas.

\begin{figure}[t]
  \centering
  \includegraphics[width=0.9\columnwidth]{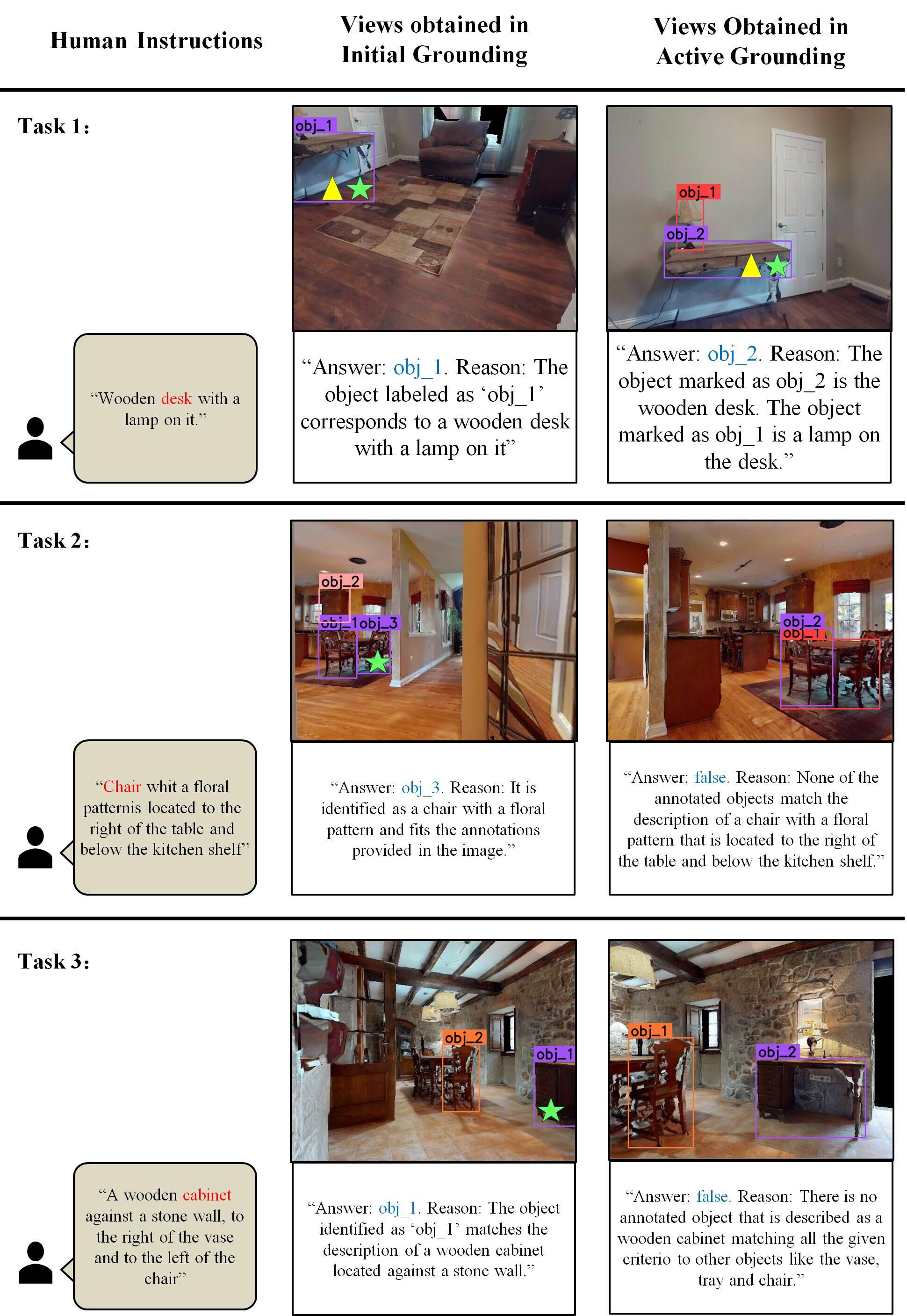}
  \caption{\textbf{Comparison of VLM grounding results.} Each task includes an instruction and annotated images used for grounding in both the initial grounding phase and the active grounding phase, along with their corresponding VLM text outputs. The target object in the instruction is highlighted in red font. The grounding results provided by the VLM are highlighted in blue font, where \textit{obj\_i} indicates that the $i$-th bounding box in the annotated image (marked with a green asterisk) is the grounding target, and \textit{false} indicates that the target object is not present in the current annotated image. If the annotated image contains the ground truth of target, it is marked with a yellow triangle.}
  \label{fig:viewpoint}
\end{figure}

We utilize a genetic algorithm implemented in the DEAP\cite{DEAP_JMLR2012} framework to tackle this multi-objective discrete optimization problem.
Figure \ref{fig:viewpoint} illustrates the different grounding results in the initial grounding and active grounding phases.
In Task 1, the robot successfully grounded the ground truth target in the initial phase and reconfirmed the grounding result in the active grounding phase.
In Tasks 2 and 3, due to the initial observation point being too far from the target and the observation angle not fully capturing the target, the robot grounded an incorrect target. However, in the active grounding phase, from a better observation position, the robot successfully eliminated the incorrect target, avoiding false negative errors.

\subsubsection{Memory-based Reserved Grounding}

During the exploration and navigation phase, active grounding can effectively mitigate initial grounding errors caused by limited perspectives. However, due to the complexity of the environment and the ambiguity of language descriptions, accurately identifying the target can still be challenging even during active grounding. In such cases, individuals typically adopt a strategy: when searching for a specific object in an unfamiliar environment, if they encounter an uncertain target, they tend to note it down first, continue exploring the environment, and then make a comprehensive judgment based on all gathered information.
We seek to equip robots with similar memory and recall capabilities, leading us to propose a memory-based reserved grounding method.

Unlike the previous grounding stages, the reserved grounding stage no longer relies on current observations but instead draws on key observations in visual memory to identify the target.
Specifically, we uniformly sample \(N_{\text{samples}}\) keyframes from all key observations in visual memory.
These annotated images \( \{I^{j}_{\text{annot}}\}_{j=1}^{N_{\text{samples}}} \) are sequentially numbered from 1 to \(N_{\text{samples}}\) and input into the Vision-Language Model (VLM) for grounding.
The output format of the VLM is predefined as a sequence number \(j\) (\(j = 1, \ldots, N_{\text{samples}}\)) of an annotated image and a numeric identifier \(i\) (\(i = 1, \ldots, N_{\text{object}}\)) of a bounding box.
Finally, the spatial coordinates of the target object can be obtained from the object memory, and the robot will navigate directly to this coordinate.

\subsection{Exploration and Navigation Policy}

If no object is grounded, we use the frontier-based exploration method described in \cite{yu2024vln} to continuously search for the goal in the environment. This strategy builds multiple similarity-value semantic maps about the goal. In order to enable the robot to navigate from the current position to the long-term goal point, we use the Fast Marching Method (FMM) \cite{sethian1996fast}. After that, the robot executes the greedy strategy at each step to give the current action \(a_t \in A\).

\section{Experiments}

In our experiments, we aim to demonstrate several aspects of our approach.
Firstly, we evaluate in simulation how effective is our method in grounding complex semantic text, particularly in comparison to current mainstream methods for language object navigation.
Secondly, through ablation studies, we determine how each module in our method contributes individually.
Finally, we deploy our framework on real-world mobile robots, especially when faced with complex language and environmental challenges.

\subsection{Simulation Setup and Implementation Details}

To validate the effectiveness of our proposed approach, we first conduct evaluations in simulation using the 3D indoor simulator, Habitat~\cite{szot2021habitat}.
The observation space comprises 480×640 RGB-D images, base odometry, and a natural language description of the target object.
The action space includes six discrete actions: \texttt{MOVE FORWARD}, \texttt{TURN LEFT}, \texttt{TURN RIGHT}, \texttt{LOOK UP}, \texttt{LOOK DOWN}, and \texttt{STOP}.
An episode is considered successful if the agent executes \texttt{STOP} within 0.3 meters of the target object and within 500 steps.

For memory construction, we employ DINO~\cite{liu2023grounding} as the open-vocabulary object detector and the Segment Anything Model (SAM)~\cite{kirillov2023segany} as the segmentation backbone.
During the Memory-Reserved Active Grounding phase, GPT-4o serves as the vision-language reasoning model.
The weights in our viewpoint optimization module are set as follows: \(w_{\text{visible}} = 15.0\), \(w_{\text{fov}} = 7.0\), \(w_{\text{distance}} = 1.0\), and \(C_{\text{infeasible}} = 1000.0\).

For historical memory replay, we cap the memory buffer at \(m_{\text{max}} = 13\). When the number of stored memories exceeds this threshold, uniform sampling is applied based on their acquisition timestamps.

\subsection{L-ZSON in Simulation}

We evaluate our method in simulation using the Habitat platform. Implementation details and parameter settings are provided in the Appendix.

\subsubsection{Dataset}

GOAT-Bench\cite{Khanna_2024_CVPR} is a comprehensive benchmark designed to evaluate visual language navigation tasks in complex indoor environments. We use the language instruction components from the GOAT-Bench validation set to assess our model's ability to understand and execute natural language instructions. The validation set includes 36 scenes and 316 language instructions covering target spatial relations and physical properties. We utilize this data to evaluate the performance of our visual language navigation framework, focusing on the effective comprehension and execution of natural language instructions.

\subsubsection{Baseline Methods}

To evaluate our method's performance on the L-ZSON task, we compared it with several techniques: CLIP-based methods \cite{radford2021learning}, CLIP on Wheels (CoW) \cite{gadre2023cows}, LLM2CLIP-based \cite{huang2024llm2clippowerfullanguagemodel}, and VLN-Game \cite{yu2024vln}. Notably, VLN-Game previously achieved SOTA performance on the GOAT-bench language instruction navigation dataset.

\begin{itemize}

  \item CLIP-based \cite{radford2021learning} methods use CLIP as the visual grounder, matching the language target description with the robot's RGB observations based on the cosine similarity scores between CLIP features.
  \item CoW \cite{gadre2023cows} adapts open-vocabulary models for Language-driven zero-shot object navigation. It processes egocentric RGB-D images and language-specified object goals, updating a top-down map with RGB-D observations and pose estimates. CoW uses an exploration policy and a zero-shot object localization module to track the target object's location. When confidence in the target's location exceeds a threshold, CoW navigates to the goal and issues a STOP action.
  \item LLM2CLIP-based \cite{huang2024llm2clippowerfullanguagemodel} enhances the foundational CLIP model by integrating large language models (LLMs) with the pretrained CLIP visual encoder. This integration leverages LLMs' advanced text understanding and open-world knowledge to improve CLIP's ability to process long and complex captions. In our experiments, we use existing methods for exploration and planning, and replace the grounding component with LLM2CLIP to evaluate its effectiveness in the L-ZSON task.
  \item VLN-Game \cite{yu2024vln} is a game-theoretic vision-language navigation framework capable of exploring unknown environments and identifying complex targets based on object language descriptions. This framework constructs a 3D object-centric spatial map by integrating pre-trained visual-language features with a 3D reconstruction of the environment, and employs a game-theoretic vision-language model to identify the target that best matches the given language description.

\end{itemize}

\subsubsection{Evaluation Metrics}

We adopt the evaluation metrics of Success Rate (SR), Success weighted by Path Length (SPL), and Distance to Goal (DTG) as proposed in \cite{anderson2018evaluation}.
SPL assesses the efficiency of the path taken by the agent. It is calculated as the ratio of the shortest possible trajectory length to the actual trajectory length, but only for successful episodes.
DTG is defined as the distance between the agent's final position and the target goal at the conclusion of each episode. It provides a measure of the proximity of the agent to the goal at the conclusion of the episode, irrespective of whether the episode was successful.

\subsubsection{Result and Discussion}
The quantitative results of the comparison study are presented in Table \ref{tab:table1}. Both the CLIP-Based and CoW methods, which use CLIP as their vision-language model (VLM), show lower performance across all metrics. This is likely due to CLIP's inherent "bag of words" nature, which struggles with complex and lengthy language descriptions, limiting its effectiveness in language-guided navigation tasks. The LLM2CLIP-based method improves CLIP's ability to understand long text descriptions by combining a language model with a pre-trained CLIP visual encoder. However, its reliance on hyperparameters like detection thresholds results in only limited performance gains. In contrast, the VLN-Game method, which integrates generative and discriminative models, demonstrates significant improvements in this task. Nevertheless, its performance is still constrained by the single-view nature of the grounding process, preventing it from fully leveraging the capabilities of the VLM. Finally, our proposed framework outperforms all baseline methods in key evaluation metrics, achieving state-of-the-art (SOTA) performance in the L-ZSON task, and demonstrating superior navigation efficiency and accuracy.

\begin{table}[ht]
  \centering
  \caption{Comparison Experiment}
  \label{tab:table1}
  \begin{tabular}{ccccc}
    \toprule
    \textbf{Method} & \textbf{VLM} & \textbf{Success} & \textbf{SPL} & \textbf{DTG} \\
    \midrule
    CLIP-Based & CLIP & 0.253 & 0.106 & 4.697 \\
    \midrule
    CoW & CLIP & 0.285 & 0.119 & 4.527 \\
    \midrule
    LLM2CLIP & LLM2CLIP & 0.313 & 0.149 & 4.847 \\
    \midrule
    \multirow{2}{*}{VLN-Game} & GPT4o-mini & 0.354 & 0.118 & 4.119 \\
    & GPT4o & 0.367 & 0.132 & 4.093 \\
    \midrule
    Ours & GPT4o & \textbf{0.408} & \textbf{0.156}  & \textbf{4.027} \\
    \bottomrule
  \end{tabular}
\end{table}

\subsubsection{Ablation study}

To evaluate the contributions of individual components in our framework, we conducted an ablation study. The results, summarized in Table \ref{tab:ablation_study}, demonstrate the performance changes when specific modules are removed.

We first evaluate the configuration where both the Active Grounding Module and the Memory Replay Strategy are removed (w/o A-M-G), and further compare two visual prompting strategies: using only annotated images (w/o A-M-G and w/o raw) versus using both annotated and raw RGB images. The results indicate that incorporating raw images improves performance in the language-guided navigation task. We hypothesize that raw RGB images provide richer semantic cues about the environment, thereby enabling the VLM to better interpret the language instructions.

We then assess the impact of removing each component individually—either the Active Grounding Module (w/o A-G) or the Memory Replay Strategy (w/o M-G). The full MAG-NAV framework consistently outperforms all ablated versions, highlighting the critical role of both the Active Grounding Module and the Memory Replay Strategy in achieving optimal performance.

\begin{table}[h!]
  \centering
  \caption{Results of Ablation Study in HM3D}
  \label{tab:ablation_study}
  \begin{tabular}{cccc}
    \toprule
    \textbf{Method} & \textbf{Success} & \textbf{SPL} & \textbf{DTG} \\
    \midrule
    w/o A-M-G and w/o raw & 0.263 & 0.136 & 5.482 \\
    w/o A-M-G & \textbf{0.285} & \textbf{0.149} & \textbf{5.137} \\
    \midrule
    w/o A-G & 0.307 & 0.142 & 5.316 \\
    w/o M-G & 0.313 & 0.159 & 5.058 \\
    MAG-NAV & \textbf{0.405} & \textbf{0.162}  & \textbf{4.342} \\
    \bottomrule
  \end{tabular}
\end{table}

\subsection{Real-World Experiments}

We deploy our proposed framework on a Go-2 quadruped robot platform equipped with a RealSense D455 RGB-D camera and a Livox Mid-360 LiDAR.
The LiDAR is used exclusively for grid map construction via the GMapping algorithm and for providing odometry information.

Using RGB-D observations and odometry, our system selects long-term navigation goals for the robot in real time. The A* algorithm is adopted for global path planning, while Model Predictive Control (MPC) is used for local planning and control execution.

To evaluate our system in real-world scenarios, we design four navigation tasks in an office-like environment. These tasks are crafted to highlight the advantages of VLM-based grounding, active viewpoint adjustment, and memory-reserved grounding. Video demonstrations of all tasks are provided in the supplementary materials.

\begin{enumerate}
    \item \textbf{Active Grounding Task.} The robot is given the instruction: “Please find my black bag on the red stool next to the trash can.” The environment contains a black bag on a red stool near a trash can and another black backpack on a red foldable bed with no trash can nearby. We compare the performance with and without the active viewpoint planning module. 

    With the active grounding enabled, the robot initially misidentifies the backpack on the foldable bed as the target. It then performs viewpoint optimization to obtain a better observation and reevaluates the scene using the VLM, successfully eliminating the false positive and navigating to the correct target. Without the active grounding module, the robot directly navigates to the incorrect backpack and terminates the episode.

    \item \textbf{VLM Grounding Task.} The instruction is: “Take the jacket hanging on the back of the black chair.” Two jackets are present in the environment: one hanging on a whiteboard and the other on the back of a black chair. We compare our VLM-grounding method with a baseline using CLIP-based grounding. Our approach first briefly examines the jacket on the whiteboard and, after interaction with the VLM, rules it out. The robot then explores further, correctly identifies the jacket on the black chair, and stops at the target. In contrast, the CLIP-based approach leads the robot directly to the incorrect jacket on the whiteboard.

    \item \textbf{Reserved Grounding Task.} The robot is sequentially given two instructions: “Find the person sitting on the sofa” and “Navigate to the black schoolbag you passed by.” The black schoolbag is located along the path toward the sofa. Upon receiving the first instruction, the robot navigates to the sofa and stops. Upon receiving the second instruction, the robot utilizes its memory-reserved grounding module to replay past visual memories and successfully localizes and navigates to the previously encountered schoolbag.
\end{enumerate}

\section{CONCLUSIONS}

In this work, we propose MAG-NAV, a training-free navigation framework designed for language-driven zero-shot object navigation(L-ZSON) tasks.
Our approach utilizes visual and object memory primitives to build and retain key memories of the agent, and employs a viewpoint-based active grounding algorithm to enhance the grounding ability of the VLM with visual prompts.
Through simulation experiments on the HM3D dataset, our method performs well in processing complex text instructions and outperforms the current state-of-the-art in the L-ZSON task.
Additionally, we deploy this framework on a real-world quadruped robot, demonstrating its ability to transfer directly from simulation to real environments, significantly improving target navigation accuracy in complex settings.
However, the grounding process based on GPT-4o reasoning may introduce latency, potentially becoming a bottleneck that limits its application in real-time robotic systems.
Despite these limitations, MAG-NAV demonstrates significant progress in visual grounding for robotic navigation and paves the way for further integration of Vision-Language Models (VLM) with robotics in the future.


\clearpage
\bibliographystyle{IEEEtran}
\bibliography{vln_references}

\end{document}